\documentclass[runningheads,a4paper]{llncs}

\usepackage{graphicx}
\usepackage{textcomp}
\usepackage{listings}
\usepackage{subfiles}
\usepackage{subfigure}
\usepackage{amsmath}
\usepackage{url}
\usepackage{enumitem}  
\usepackage[utf8]{inputenc}
\usepackage{csquotes}
\usepackage{fancyhdr}
\usepackage{float}
\usepackage[T1]{fontenc}
\usepackage[utf8]{inputenc}
\usepackage{authblk}

\begin{document}

\pagestyle{headings}  % switches on printing of running heads

\title{Image Processing Based Scene-Text Detection and Recognition with Tesseract}

\titlerunning{Scene Text Recognition using Image processing}  % abbreviated title (for running head)

\author{Ebin Zacharias\inst{1}, Martin Teuchler\inst{2} \and B\'en\'edicte Bernier\inst{3}}

\authorrunning{Zacharias et al.} % abbreviated author list (for running head)
\institute{\email{$^1$ebinzacharias@gmail.com}, \email{$^2$Martin.Teuchler@kamag.com}, and \email{$^3$Benedicte.Bernier@kamag.com}}

\maketitle              % typeset the title of the contribution

\begin{abstract}
Text Recognition is one of the challenging tasks of computer vision with considerable practical interest. Optical character recognition (OCR) enables different applications for automation. This project focuses on word detection and recognition in natural images. In comparison to reading text in scanned documents, the targeted problem is significantly more challenging. The use case in focus facilitates the possibility to detect the text area in natural scenes with greater accuracy because of the availability of images under constraints. This is achieved using a camera mounted on a truck capturing likewise images round-the-clock. The detected text area is then recognized using Tesseract OCR engine. Even though it benefits low computational power requirements, the model is limited to only specific use cases. This paper discusses a critical false-positive case scenario occurred while testing and elaborates the strategy used to alleviate the problem. The project achieved a correct character recognition rate of more than 80\%. This paper outlines the stages of development, the major challenges and some of the interesting findings of the project. 
\end{abstract}

% This is your content. Name the sections appropriately
\section{Introduction}

Text recognition is an interesting and highly researched field of computer vision. It encourages the applications of automation and hence reducing human efforts. Optical Character Recognition (OCR) works efficiently with printed text documents \cite{ramesh2018improving}. However, scene text recognition is a hard task and has only recently gained attention from the computer vision community. The accuracy of OCR predominantly depends on image preprocessing and quality of the image. Moreover, factors such as font size, geometric distortions and illumination effects are critical in recognizing the required text accurately. This project aims at implementing a scene text recognition system for industrial application.

The main objective is to extract ILU codes (Intermodal Loading Units) printed on the rear end of the swap bodies. ILU code is a unique number consisting of mainly three parts. The first four characters represent the owner-key which is issued by the Union for Road-Rail Combined Transport (UIRR) followed by a 6 digit registration number which can be freely allocated by the owner. The last number is a check digit which is obtained by a defined calculation procedure \cite{website1}. Figure 1 is an example of the ILU code which is the main focus of this project and shows the pattern of text that is to be translated. The project is achieved in two steps and utilizes simple image processing techniques to detect the text region in the image. Clearly, the characters in the ILU code has a pattern and can be verified using the unique checksum calculation. Tesseract 5 is used for text recognition which is a deep learning-based model and utilizes LSTM (Long Short Term Memory). However, testing on a larger dataset resulted in notable false-positive scenarios. The checksum digits were altered corresponding to the wrong detection of the registration number during text recognition using tesseract. The steps taken to mitigate this problem is explained in this paper. Finally, this paper discusses the test results and possible future applications.

\begin{figure}[H]
\centering
\includegraphics[width=0.5\linewidth]{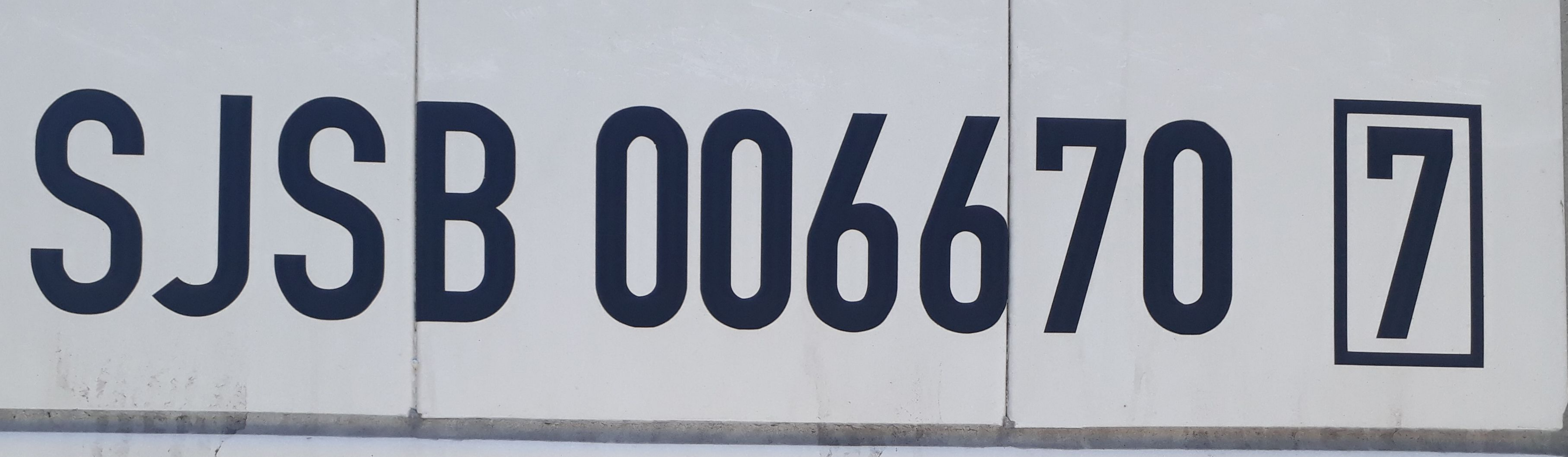}
\caption{A sample image of the ILU code.}
\label{fig:verticalcell}
\end{figure}

\section{Methodology}

This paper proposes a text recognition pipeline for translating text in the natural scenes. The goal is achieved in two steps. Firstly, the text area is detected which plays an important role in the overall performance of the OCR engine. Secondly, the detected text area is translated using tesseract V5. Tesseract is an accurate and open-sourced OCR engine from Google \cite{zhang2010mobile}. Figure 2 is a flowchart depicting the major steps involved in text recognition. A detailed description of the steps involved is discussed in the following sections.

\begin{figure}[H]
\centering
\includegraphics[width=0.7\linewidth]{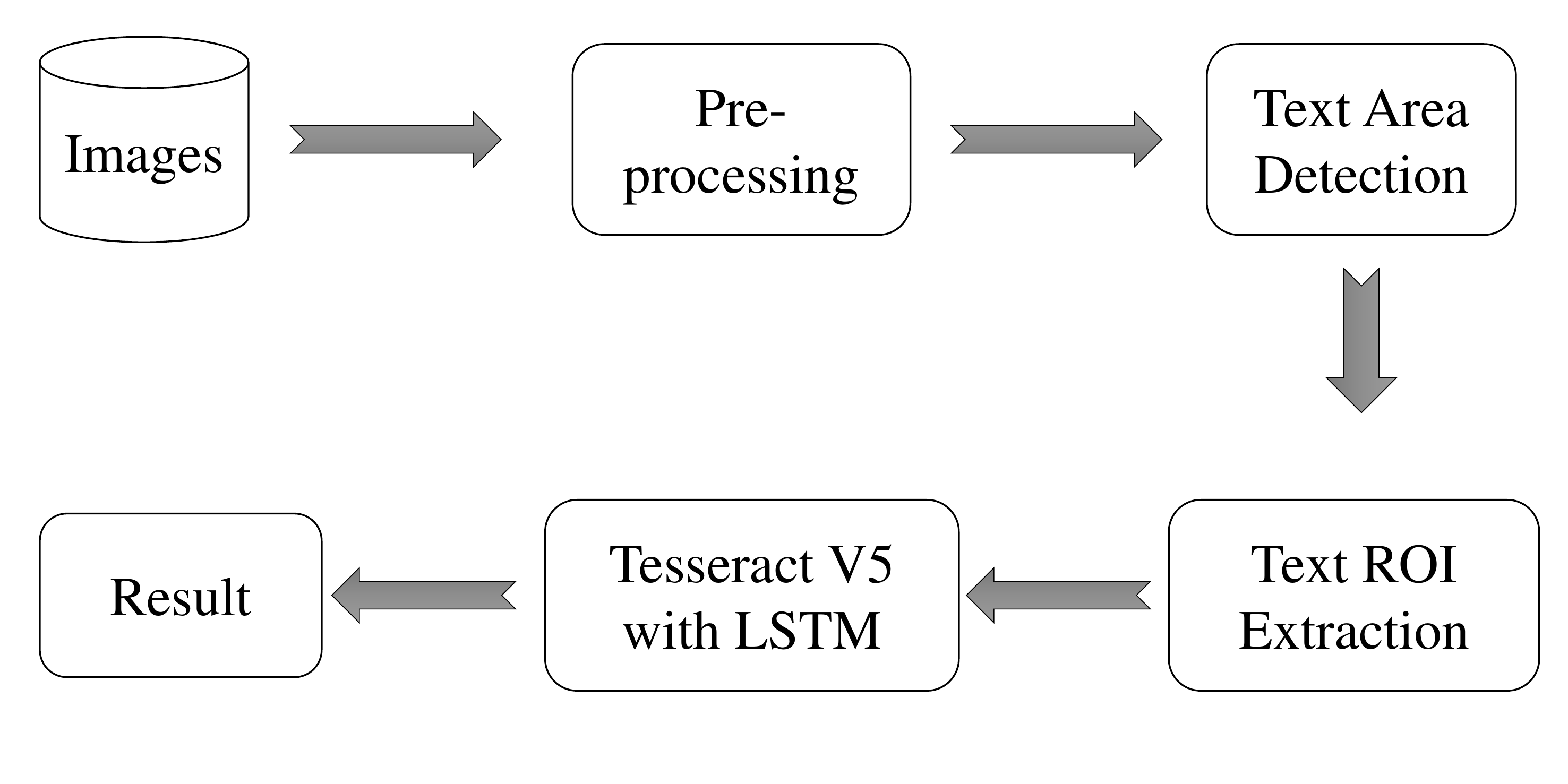}
\caption{Steps involved in detecting the text recognition.}
\label{fig:verticalcell}
\end{figure}

\subsection{Data Collection}

The images are captured using a USB camera module mounted at the rear end of a logistic truck. Images are captured when the swap body is picked up, transported and set down at the target location. Furthermore, all images are stored in the cloud using Microsoft Azure. The images obtained contains noise and it is necessary to pre-process the images. Besides, images are captured round-the-clock and hence there is a significant variation in the brightness of images.      

\subsection{Text Detection}

\subsubsection{Preprocessing Steps}

Preprocessing is a key component and undoubtedly affects the results in successive steps \cite{karthick2019steps}. Images stored in the cloud contains a significant amount of noise. The camera is mounted at a slightly inclined angle with respect to the text required for recognition. The following methods are used to achieve the preprocessing stage which has proved critical in text extraction. Apart from the low illumination of night images, there is a reverse flashlight falling exactly on the checksum number which was a challenge for obtaining better results. Fixing the illumination of the image is an important requirement for the Tesseract OCR engine to work efficiently. Figure 3 shows a sample of the night image used.

\begin{figure}[H]
\centering
\includegraphics[width=0.8\linewidth]{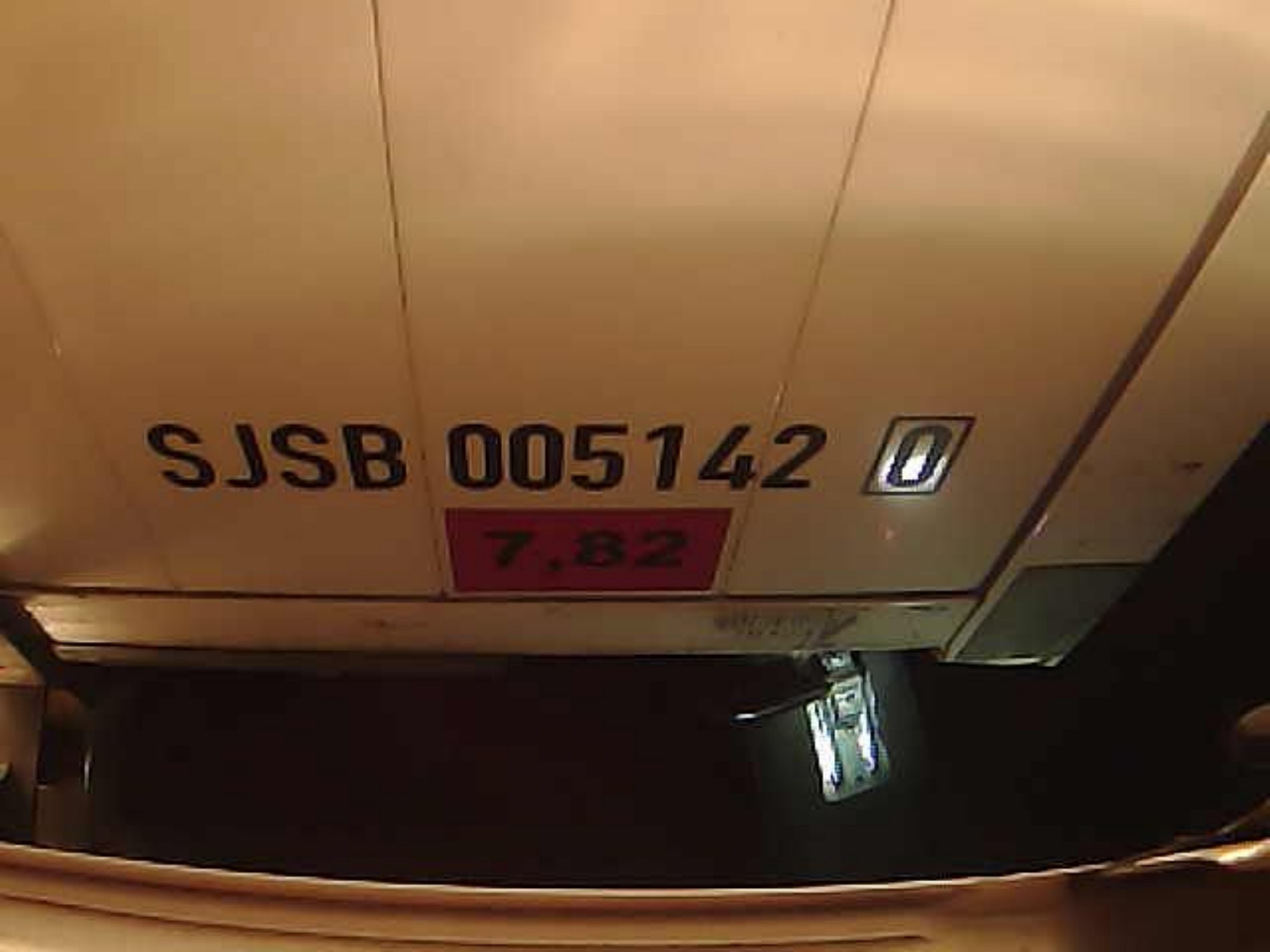}
\caption{Sample of image at night.}
\label{fig:verticalcell}
\end{figure}

\paragraph{Image cropping:} The images obtained are constrained and the text area of interest is always on the lower right side of the image. The constraints and uniqueness of input images helped in fixing a cropping area so that the unwanted areas from the image were eliminated. Accordingly, the images are cropped based on a predefined crop area.

\paragraph{Brightness check and gamma correction:} After cropping the images, the next problem was the varying brightness levels of the image. The night images also faced the problem of an additional flash-light which markedly changed the illumination effects on the image. Methodically to differentiate the images, the brightness of the images were calculated using root mean square (RMS) pixel brightness method. The images are then classified into different categories followed by gamma correction. This approach demonstrated success when the gamma values of the images are varied depending on the brightness levels of the is. A lookup table (LUT) was build mapping the input pixel values to the output gamma-corrected values. It resulted in faster gamma correction using OpenCV.

\paragraph{Skew correction:} As already discussed above, the images are captured at an angle and to deal with the geometric distortion caused by the camera position, a skew correction was performed to get an aligned text area. In addition, deskewing and dewarping the image are seen important for efficient text recognition in later stages. After finding the edges using a canny edge detector, the strong lines in the image are detected using hough line transformation. Progressive probabilistic hough line transformation is used which reduces the computation effort \cite{matas2000robust}. There are two main arguments: minimum length of line and the maximum allowed gap between line segments which together helped to detect the strong lengthy lines in the image. Finally, geometric transformation such as perspective transformation is used to achieve necessary skew correction. The transformation matrix for perspective transformation is achieved from above-mentioned hough line detection.

\subsubsection{Finding the Text Area}

In order to reduce the noise, Gaussian blurring is applied to the image. Using a blackhat operator, dark regions in the images are revealed. The text in the images is dark and against a light background which reinforced the blackhat operation. As a next step, using a Scharr operator, the gradient magnitude representation of the blackhat image is computed. Scharr operator optimises the rotation symmetry by minimizing the weighted mean squared angular error \cite{website2}. The gaps between the characters are closed using a morphological closing operation. Finally, the Otsu thresholding method is applied to the image.

\subsubsection{Contour Detection and Region of Interest (ROI)}

The contours of the detected area are computed and the contours are sorted based on their size. This step is helpful in extracting the text area considering its larger size in the image. The bounding box of the contour is computed and the ROI is extracted if the aspect ratio and the width of the bounding box are within an acceptable criterion. Figure 4 illustrates the steps involved in detecting the text area. Tesseract has a reduced accuracy when the character height is less than 20 pixels \cite{ramesh2018improving}. In order to overcome this flaw, the final ROI is resized accordingly.

\begin{figure}[H]
\centering
\includegraphics[width=0.9\linewidth]{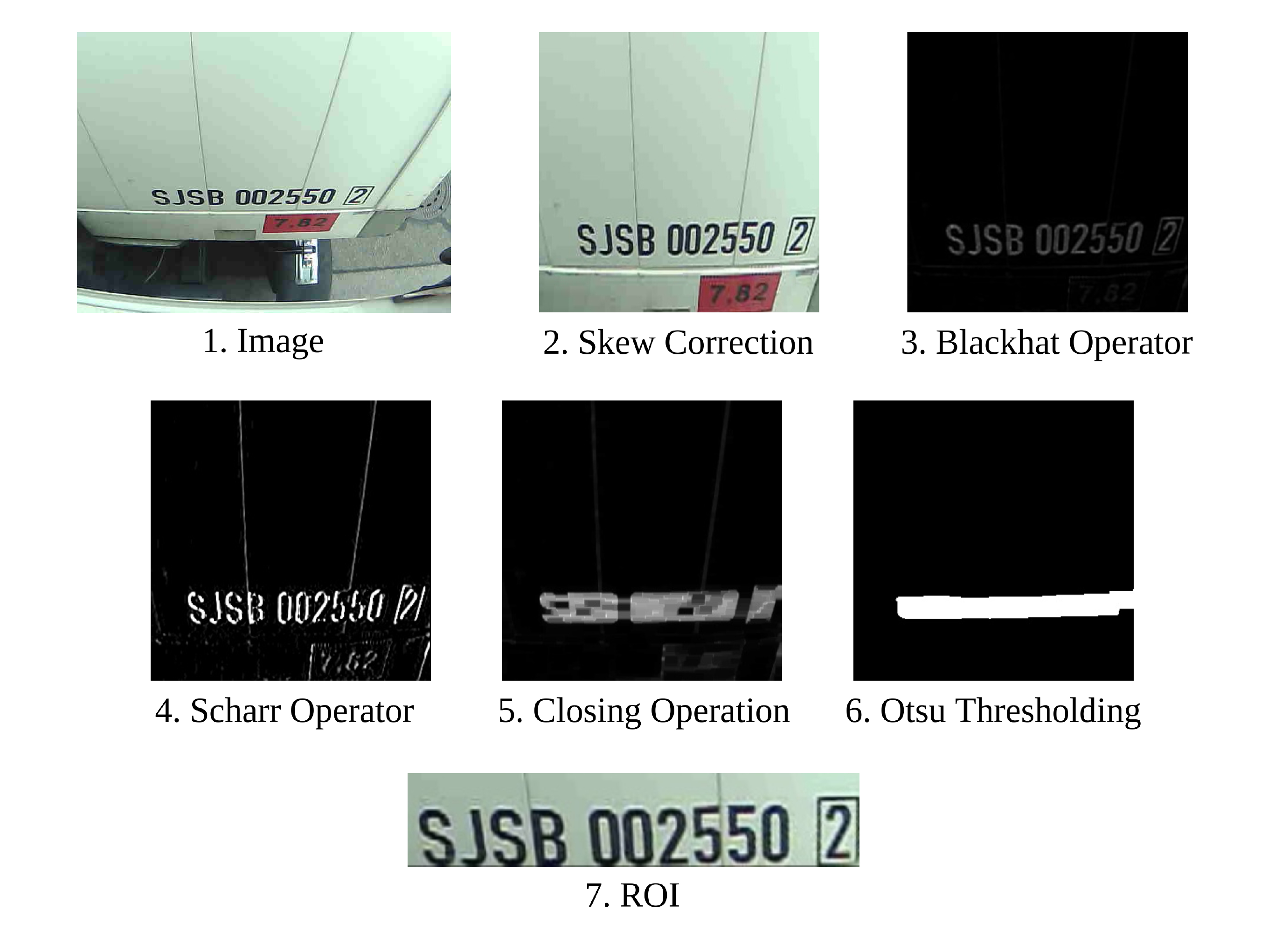}
\caption{Steps involved in detecting the text area.}
\label{fig:verticalcell}
\end{figure}

\subsection{Text Recognition}

Text recognition is performed using Tesseract V5 from Google. Tesseract is a deep learning-based text recognition model which not only possess high accuracy levels but also supports a wide variety of Languages. However, Tesseract OCR consists of some important assumptions for text recognition. It performs accurately with documented text and the accuracy is generally limited to controlled conditions \cite{website3}. Hence, preprocessing of images and extraction of the text area are performed keeping these factors in mind.

This project used Tesseract 5 with LSTM (Long Short Term Memory) as the OCR engine mode. Furthermore, fine-tuning the page segmentation modes (PSM) resulted in significant improvements in the overall accuracy of the pipeline. After careful tests and comparisons, two different page segmentation methods were used. For the first part of recognition the image was treated as a single character and for the second part, the automatic page segmentation method was utilized. A detailed description of the steps involved is explained in the next section.

The detected text is then evaluated using text post-processing. The checksum value is separately calculated each time based on the first 10 characters after recognition. It is based on a given calculation procedure \cite{website1}. The last digit in square brackets is the checksum number. In order to verify the recognized text, the calculated checksum number is compared with the detected checksum number. If they are equal, then the extracted text is verified and stored as true detection. The ROI image was treated as a single character which corresponds to tesseract PSM 10.

Interestingly, later tests with a larger dataset resulted in false-positive cases which are discussed in the following section. After careful evaluation of some of the false-positive cases, the idea was to cross-verify by breaking the detected ROI and separately extract the text. A two step evaluation process was performed. The detected ROI is manually split into two parts: the registration number and the checksum number. Later, the text recognition process was performed separately on each part and cross verified with the checksum number. PSM 10 was used for recognising part 1 whereas PSM 3 was used for extracting the checksum number. PSM 3 corresponds to automatic page segmentation method and the presence of square boxes around the checksum number was a reason to try this technique. Moreover, the character was limited to only digits using tesseract arguments. This approach proved rewarding in reducing false-positive cases.

\section{Testing and Results}

The text recognition system was tested using the NVIDIA Jetson TX2 Developer Kit. Initially, a dataset of 3000 images was used for testing. After text post-processing and verifying the checksum based on known calculation, the overall character accuracy was 89.7\%. However, the accuracy was limited as further evaluation on the test result revealed a significant number of false-positive cases. The accuracy significantly dropped to 49\% after considering the false-positive cases.

The analysis showed that there were cases in which if one of the numbers in the registration number was detected wrong, the checksum number was also automatically altered and detected wrong. Notably, when the checksum number is calculated in the text post-processing step, the altered checksum number matches with the wrongly detected registration number. A high number of false-positive cases annihilate the chances of an accidental occurrence. As an attempt to eradicate this critical problem and to evaluate the results, the ROI after text area detection was split into two parts. Text recognition was performed separately as explained in the previous section and cross verified with the calculated checksum number. This approach eliminated false-positive cases. The pipeline was then tested on 7450 images and resulted in an overall accuracy of 83\%.

\section{Conclusion and Future Work}

Optical character recognition is a highly researched topic. Text recognition in a natural scene is a difficult task. This paper discussed a use case specific application with constrained images as input. The overall accuracy of 83\% was achieved which can be slightly improved with the quality of the input images. However, text recognition is not flawless when subjected to complex environmental situations. Small variations in the illumination and text angles resulted in large errors while recognizing the text. Even though Tesseract performs well with the documented text, the accuracy largely fluctuated when scene text images were tested. Surprisingly, wrong text detection however matching with the calculated number is a serious concern to examine. Training and implementing a deep learning-based model for extracting scene text images can be promising. Clearly, there is a pattern and dependability between the characters used in this project. Auto-generation of checksum number was an interesting occurrence but a critical event. The deep learning-based approach can be helpful to evaluate the dependability of such patterns of text in an image and also to analyze the influence of font type in text recognition.

%

%
% ---- Bibliography ----
%
\bibliographystyle{plain}
\bibliography{references}

\end{document}